\documentclass[11pt]{article}
\usepackage{coling2018}
\usepackage{times}
\usepackage{url}
\usepackage{latexsym}

\usepackage{amsmath}
\usepackage{amssymb}
\usepackage{enumitem}
\usepackage{fancyvrb}
\usepackage{natbib}
\usepackage{graphicx}
\usepackage{url}
\usepackage{multirow}
\usepackage{float}

\newcommand{\href}{\url}

\title{Cross-Discourse and Multilingual Exploration of Textual Corpora with
the DualNeighbors Algorithm}

\author{Taylor Arnold \\
  University of Richmond \\
  Mathematics and Computer Science \\
  28 Westhampton Way \\
  Richmond, VA, USA \\
  {\tt tarnold2@richmond.edu} \\\And
  Lauren Tilton \\
  University of Richmond \\
  Rhetoric and Communication Studies \\
  28 Westhampton Way \\
  Richmond, VA, USA \\
  {\tt ltilton@richmond.edu} \\}

\date{20 June 2018}

\begin{document}
\maketitle
\begin{abstract}
Word choice is dependent on the cultural context of writers and their
subjects. Different words are used to describe similar actions, objects,
and features based on factors such as class, race, gender, geography
and political affinity. Exploratory techniques based on locating and
counting words may, therefore, lead to conclusions that reinforce
culturally inflected boundaries. We offer a new method, the DualNeighbors
algorithm, for linking thematically similar documents both
within and across discursive and linguistic barriers to reveal cross-cultural
connections. Qualitative and
quantitative evaluations of this technique are shown as applied to two
cultural datasets of interest to researchers across the humanities and
social sciences. An open-source implementation of the DualNeighbors algorithm
is provided to assist in its application.
\end{abstract}

\section{Introduction}

\blfootnote{
    \hspace{-0.65cm}  
    This work is licensed under a Creative Commons
    Attribution 4.0 International License.
    License details:
    \url{http://creativecommons.org/licenses/by/4.0/}
}

Text analysis is aided by a wide range of tools
and techniques for detecting and locating themes and subjects.
Key words in context (KWiC), for example, is a method from corpus
linguistics for extracting short snippets of text containing a predefined
set of words \citep{luhn1960key, gries2009quantitative}. Systems for full
text queries have been implemented by
institutions such as the Library of Congress, the Social Science Research
Network, and the Internet Archive \citep{brenton2016search}.
As demonstrated by the centrality of search engines to the
internet, word-based search algorithms are powerful tools for locating
relevant information within a large body of textual data.

Exploring a collection of materials by searching for words poses a potential
issue. Language is known to be highly dependent on the cultural factors
that shape both the writer and subject matter. As concisely described by
\citet{foucault1969archeologie}, ``We know perfectly well that we
are not free to say just anything, that we cannot simply speak of anything,
when we like or where we like; not just anyone, finally, may speak of just
anything.'' Searching through a corpus by words and phrases reveals a
particular discourse or sub-theme but can make it challenging to identify a
broader picture. Collections with multilingual data pose an extreme form of
this challenge, with the potential for important portions of a large corpus to
go without notice when using traditional search techniques.


\begin{table*}
  \centering
  \includegraphics[width=0.9\linewidth]{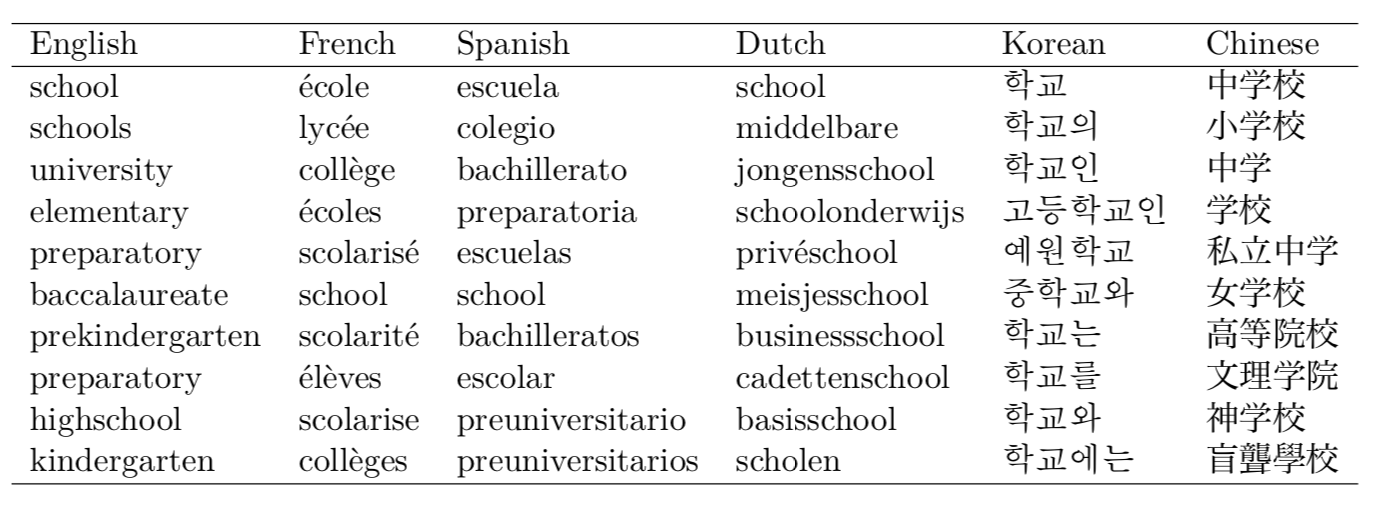}
  \caption{Nearest neighbors of the English word ``school'' in a multilingual
embedding space.}
  \label{tab:embed}
\end{table*}

Our works build off of recent research in word embeddings to provide a novel
exploratory recommender system that ensures recommendations can cut across
discursive and linguistic boundaries. We define two
similarity measurements on a corpus: one based on word usage and another
based on multilingual word embeddings. For any document in the corpus, our
DualNeighbors algorithm returns the nearest neighbors from each of these two
similarity measurements. Iteratively following recommendations through the
corpus provides a coherent way of understanding structures and patterns within
the data.

The remainder of this article is organized as follows.
In Section~\ref{sec:prior} we first give a brief overview of prior work in
the field of word embeddings, recommender systems, and multilingual search.
We then provide a concise motivation and algorithmic description of the
DualNeighbors algorithm in Sections~\ref{sec:goal} and \ref{sec:method}.
Next, we qualitatively (Section~\ref{sec:qual}) and quantitatively
(Section~\ref{sec:quant}) assess the algorithm as applied to (i) a large
collection of captions from an iconic archive of American photography, and
(ii) a collection of multilingual Twitter news feeds. Finally, we conclude
with a brief description of the implementation of our algorithm.

\section{Related Work} \label{sec:prior}

\subsection{Word Embeddings} \label{ssec:embed}


Given a lexicon of terms $L$, a word embedding is a function that maps each
term into a $p$-dimensional sequence of numbers \citep{mikolov2013distributed}.
The embedding implicitly describes relationships between words, with similar
terms being projected into similar sequences of numbers
\citep{goldberg2014word2vec}. Word embeddings are typically derived by placing
them as the first layer of a neural network and updating the embeddings by a
supervised learning task \citep{joulin2017bag}. General purpose embeddings can
be constructed by using a generic training task, such as predicting a word as a
function of its neighbors, over a large corpus \citep{mikolov2013efficient}.
These embeddings can be distributed and used as an input to other text
processing tasks. For example, the pre-trained fastText embeddings provide
$300$-dimensional word embeddings for $157$ languages \citep{grave2018learning}.

While there is meaningful information in the distances between words in an
embedding space, there is no particular significance attached to each of its
dimensions.
Recent work has drawn on this degree of freedom to show that two independently
trained word embeddings can be aligned by rotating one embedding to match
another. When two embeddings from different languages are aligned, by way of
matching a small set of manual translations, it is possible to embed a
multilingual lexicon into a common space \citep{smith2017offline}.
Table~\ref{tab:embed} shows the nearest word neighbors to the English term
`school' in six different languages. The closest neighbor in each language is
an approximate translation of the term; other neighbors include particular
types of schools and different word forms of the base term.

\subsection{Word Embedding Recommendations} \label{ssec:rec}

The ability of word embeddings to capture semantic similarities make them an
excellent choice for improving query and recommendation systems. The word
movers distance of \citet{kusner2015word} uses embeddings to describe a new
document similarity metric and \citet{li2016topic} uses them to extend topic
models to corpora containing very short texts. Works by \citet{ozsoy2016word}
and \citet{manotumruksa2016modelling} utilize word embeddings as additional
features within a larger supervised learning task. Others have, rather than
using pre-trained word embeddings, developed techniques for learning item
embeddings directly from a training corpus
\citep{barkan2016item2vec, vasile2016meta, biswas2017mrnet}.

Our approach most closely builds off of the query expansion techniques
of \citet{zamani2016estimating} and \citet{de2016representation}. In both
papers, the words found in the source document are combined with other terms
that are close within the embedding space. Similarity metrics are then derived
using standard probabilistic and distance-based methods, respectively.
Both methods are evaluated by comparing the recommendations to observed user
behavior.

\subsection{Multilingual Cultural Heritage Data} \label{ssec:multi}

Indexing and linking multilingual cultural heritage data is an important and
active area of research. Much of the prior work on this task has focused on the
use of semantic enrichment and linked open data, specifically through named
entity recognition (NER). Named entities are often written similarly across
languages, making them relatively easy points of reference to link across
multilingual datasets \citep{pappu2017lightweight}. \citet{de2017semantic}
recently developed MERCKX, a system for combining NER and DBpedia for the
semantic enrichment of multilingual archive records, built off of a
multilingual extension of DBpedia Spotlight \citep{daiber2013improving}. To the
best of our knowledge, multilingual word embeddings have not been previously
adapted to the exploration of cultural heritage datasets.

\section{Goal and Approach} \label{sec:goal}

Our goal is to define an algorithm that takes a starting document within a
corpus of texts and recommends a small set of thematically or stylistically
similar documents. One can apply this algorithm to a particular text of
interest, select one of the recommendations, and then
re-apply the algorithm to derive a new set of document suggestions. Following
this process iteratively yields a method for exploring and understanding a
textual corpus. Ideally, the collection of recommendations should be
sufficiently diverse to avoid getting stuck in a particular subset of the
corpus.


Our approach to producing document recommendations, the DualNeighbors
algorithm, constructs two distinct similarity measurements over the
corpus and returns a fixed number of closest neighbors from each
similarity method. The first measurement uses a standard TF-IDF
(term-frequency, inverse
document frequency) matrix along with cosine similarity. We call the
nearest neighbors from this set the \textit{word neighbors}; these assure that
the recommendations include texts that are very similar and relevant to the
starting document. In the second metric we replace terms in the search document
by their closest $M$ other terms within a word embedding space. The transformed
document is again compared to the rest of the corpus through TF-IDF and
cosine similarity. The resulting \textit{embedded neighbors} allow for an
increased degree of diversity and connectivity within the set of
recommendations. For example, using Table~\ref{tab:embed}, the embedding
neighbors for a document using the term ``school'' could include texts
referencing a ``university'' or ``kindergarten''.

The DualNeighbors algorithm features two crucial differences compared to other
word-embedding based query expansion techniques. Splitting the search
explicitly into two types of neighbors allows for a better balance between the
connectivity and diversity of the recommended documents.
Also, replacing the document with its closest word embeddings, rather than
augmenting
as other approaches have done, significantly improves the diversity of the
recommended documents. Additionally, by varying the number of neighbors
displayed by each method, users can manually adjust the balance between
diversity and relevance in the results. The effect of these distinctive
differences are evaluated in Table~\ref{tab:metrics} and
Section~\ref{sec:quant}.


\section{The DualNeighbors Algorithm} \label{sec:method}

Here, we provide a precise algorithmic formulation of the DualNeighbors
algorithm.
We begin with a pre-determined lexicon $L$ of
lemmatized word forms. For simplicity of notation we will assume that words are
tagged with their language, so that the English word
``fruit'' and French word ``fruit'' are distinct. Next, we take a
(possibly multilingual) $p$-dimensional word embedding function,
as in Section~\ref{ssec:embed}. For a fixed neighborhood size $M$, we can
define the neighborhood function as a function $f$ that maps each term in
$L$ to a set of new terms in the lexicon by associating each word in $L$ with
its $M$ closest (Euclidiean) neighbors. The DualNeighbors algorithm is then
given by:

\begin{enumerate}[wide, labelwidth=!, labelindent=0pt]
\item \textbf{Inputs}: A textual corpus $C$, document index
of interest $\tilde{i}$, a lexicon $L$, word neighbor function
$f$, and desired number word neighbors $N_{w}$ and embedded neighbors
$N_{e}$ to return.
\item First, apply tokenization, lemmatization, and part-of-speech
tagging models to each element in the input corpus $C$. Filter the
word forms to those found in the set $L$. Then write the corpus $C$ as
\begin{align}
C &= \{ c_i \}_{i=1}^n, \quad c_i = \{ w_{i,k_i} \}_{k_i},
\quad w_{i,k_i} \in L, \quad 1 \leq k_i \leq |L|
\end{align}
\item For each document $i$ and element $j$ in the lexicon,
compute the $n \times|L|$ dimensional binary term frequency
matrix $Y$ and TF-IDF matrix $X$ according to
\begin{align}
Y_{i, j} &= \begin{cases} 1, & l_j \in c_i \\ 0, & \text{else} \end{cases}
&X_{i, j} = Y_{i, j} \times \log \frac{n}{\sum_i Y_{i, j}}.
\end{align}
\item Simlarly, compute the embedded corpus $E$ as
\begin{align}
E &= \{ e_i \}, \quad
e_i = \bigcup_{k_i} f(w_{k_i}). \label{expcorp}
\end{align}
Define the the embedded binary term frequency matrix $Y^{emb}$
and TF-IDF matrix $X^{emb}$ as
\begin{align}
Y^{emb}_{i, j} &= \begin{cases} 1, & l_i \in e_i \\ 0, & \text{else} \end{cases}
&X^{emb}_{i, j} &= Y^{emb}_{i, j} \times \log \frac{n}{\sum_i Y_{i, j}}.
\end{align}
\item Compute the $n\times n$ document similarity matrices $S$ and $S^{emb}$
using cosine similarity, for $i \neq i'$, as
\begin{align}
S_{i, i'} &= X_{i'} X^t_i / \sqrt{X^t_i X_i}
&S^{emb}_{i, i'} = X^{emb}_{i'} X^t_i / \sqrt{X^t_i X_i},
\end{align}
where $X_i$ is the $i$th row vector of the matrix $X$ and
$S_{i, i}$ and $S^{emb}_{i, i}$ are both set to zero.
\item \textbf{Output}: The recommended documents associated with
document $\tilde{i}$ are given by:
\begin{align}
\texttt{TopN}\left(N_{w}, S_{i, {\tilde{i}}}\right) \bigcup
\texttt{TopN}\left(N_{e}, S^{emb}_{i, {\tilde{i}}}\right)
\end{align}
where $\texttt{TopN}(k, x)$ returns the indices of the largest $k$
values of $x$.
\end{enumerate}

In practice, we typically start with Step~2 of the algorithm to
determine an appropriate lexicon $L$ and cache the similarity
matrices $S$ and $S^{emb}$ for the next query. In implementation and
examples, the multilingual fastText word embeddings of
\citep{grave2018learning} used. Details of the implementation of the
algorithm are given in Section~\ref{sec:imp}.

\begin{figure*}[t!]
    \centering
    \includegraphics[width=\textwidth]{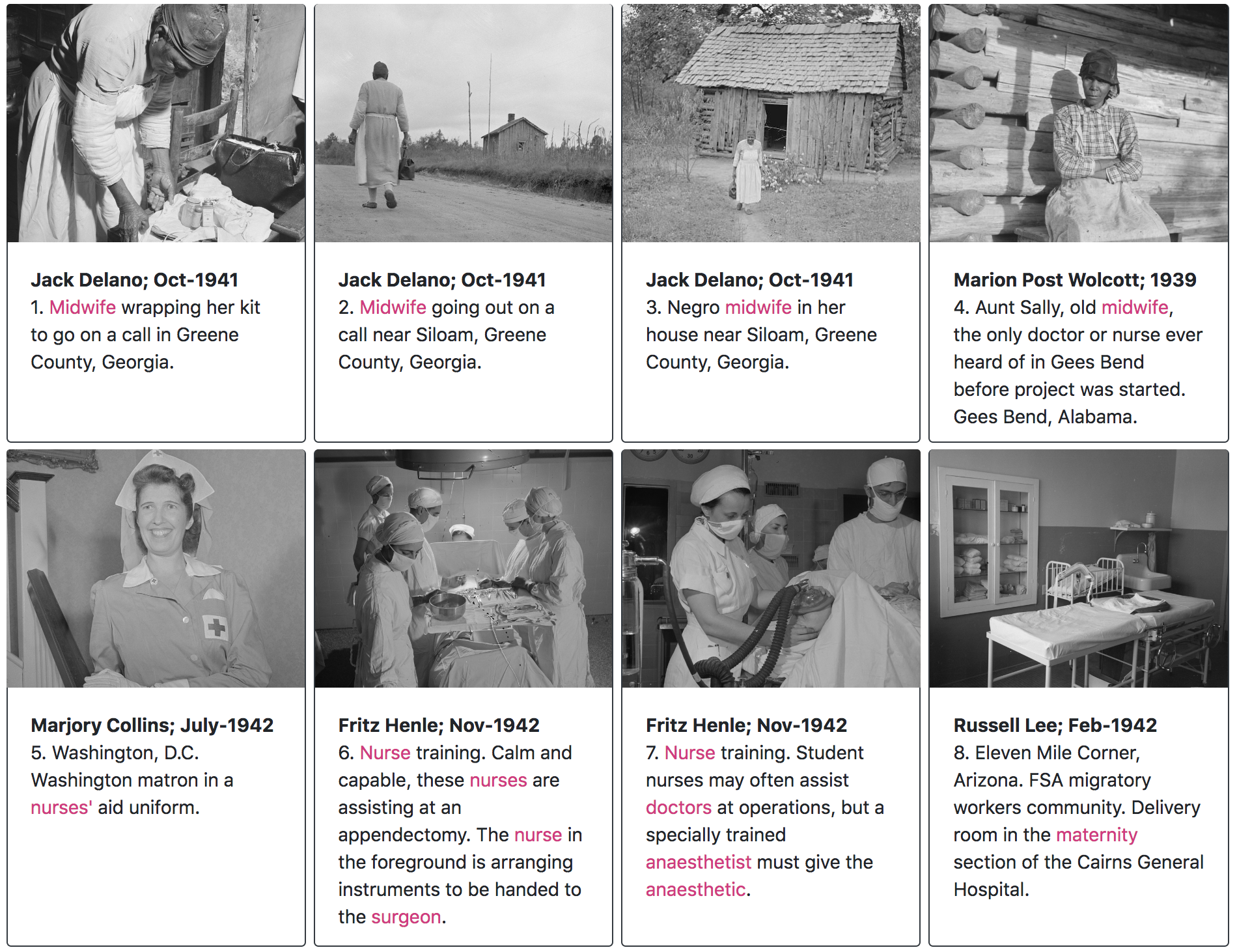}
    \caption{Example visualization of the DualNeighbors algorithm. Item 1 is
    the starting point, items 2-4 are the first three word neighbors, and
    5-8 are the first four embedding neighbors.}
    \label{fig:embed}
\end{figure*}

\section{Qualitative Evaluation} \label{sec:qual}

\begin{table*}[t!]
\centering
\begin{tabular}{p{4cm}|p{5.3cm}|p{5.3cm}}
Caption & Top-3 Word Neighbors & Top-3 Embedding Neighbors \\
  \hline
  \hline
  Grading and bunching \textbf{carrots} in the \textbf{field}. Yuma County, Arizona & $\bullet$ Bunching \textbf{carrots} in the \textbf{field}. Yuma County, Arizona & $\bullet$ Roadside display of \textbf{pumpkins} and \textbf{turnips} and other vegetables near Berlin, Connecticut \\
   & $\bullet$ Bunching \textbf{carrots}. Imperial County, California & $\bullet$ Hartford, Connecticut... Mrs. Komorosky \textbf{picking} \textbf{cucumber}s \\
   & $\bullet$ Bunching \textbf{carrots}, Edinburg, Texas & $\bullet$ Pumpkins and \textbf{turnip}s near Berlin, Connecticut \\ \hline
  Brownsville, TX. Charro Days \textbf{fiesta}. \textbf{Children}. & $\bullet$ Brownsville, Texas. Charro Days \textbf{fiesta}. & $\bullet$ \textbf{Picnic} lunch at May Day-Health Day festivities... \\
   & $\bullet$ Visitor to Taos \textbf{fiesta}, New Mexico & $\bullet$ Spectators at \textbf{childrens} races, Labor Day \textbf{celebration} ... \\
   & $\bullet$ Bingo at \textbf{fiesta}, Taos, New Mexico & $\bullet$ Detroit, Michigan. Child in \textbf{toddler} go-cart \\ \hline \hline
Imperial Brands \textbf{shareholders} revolt over \textbf{CEO}'s pay rise & $\bullet$ Evening Standard urged to declare Osborne's job with Uber \textbf{shareholder} & $\bullet$ Bruno Le Maire à Wall Street pour attirer les \textbf{investisseur}s ... \\
   & $\bullet$ Uber \textbf{CEO} Travis Kalanick should have gone years ago & $\bullet$ Pierre Bergé : Le Monde perd l'un de ses \textbf{actionnaires} \\
   & $\bullet$ \textsterling37bn paid to \textbf{shareholders} should have been invested & $\bullet$ Le pacte d'\textbf{actionnaires} de STX France en question \\ \hline
Cannes 2017: Eva Green and Joaquin Phoenix on
& $\bullet$ Five looks to know about from the SAG \textbf{red} \textbf{carpet} & $\bullet$ Festival de Cannes 2017: Bella Hadid, \textbf{rouge} \'{e}carlate sur le \textbf{tapis} \\
the \textbf{red} \textbf{carpet}  & $\bullet$ Baftas 2017: the best of the \textbf{red} \textbf{carpet} fashion& $\bullet$ \`{A} New York, \textbf{tapis} \textbf{rouge} pour Kermit la grenouille \\
& $\bullet$ Emmys 2016 fashion: the best looks on the \textbf{red} \textbf{carpet} & $\bullet$ Sur \textbf{tapis} \textbf{rouge} \\
\end{tabular}
\caption{Two FSA-OWI captions and two tweets from the \textit{Guardian}
versus \textit{Le Figaro} corpora along with the top-3 word and embedding
neighbors.}
\label{tab:captions}
\end{table*}

\subsection{FSA-OWI Captions}

Our first example applies the DualNeighbors algorithm to a corpus of
captions attached to approximately ninety thousand photographs taken between
1935 and 1943 by the U.S. Federal Government through the Farm Security
Administration and Office of War Information \citep{baldwin1968poverty}.
The collection remains one of the most historically important archives of
American photography \citep{trachtenberg1990reading}.
The majority of captions consist of a short sentence describing the scene
captured by the photographer. Photographic captions mostly come
from notes taken by individual photographers; the style and lexicon is
substantially variable across the corpus.

An example of the connections this method gives
are shown in Table~\ref{tab:captions}. For example, the word neighbors of the
caption about the farming of carrots consists of other captions related to
carrots. The embedding neighbors link to captions describing other vegetables,
including pumpkins, cucumbers and turnips. Because of the correlation between
crop types and geography, the embedding neighbors allow the search to extend
beyond the U.S. South into the Northeast. Similarly, the caption about
fiestas (a Spanish term often used to describe events in Hispanic/Latino
communities) becomes linked to similar festivals in other locations by way of
its embedding neighbors. By also including a set of word neighbors, we
additionally see other examples of events within various communities
across the Southwestern U.S..

Figure~\ref{fig:embed} shows the images along with the captions for
a particular starting document. In the first row, the word neighbors show
depictions of two older African American midwives, one in rural Georgia by
Jack Delano in 1941 and another by Marion Post Walcott in 1939. The second row
contains captions and images of embedding neighbors. Among these are two Fritz
Henle photographs of white nurses training to assist with an appendectomy,
taken in New York City in 1943. These show the practice of medicine in the
U.S. from two different perspectives.
Using only straightforward TF-IDF methods, there would otherwise have been no
obvious link between these two groups of images. The two sets were taken over a
year apart by different photographers in different cities. None of the key
terms in the two captions match each other. It would be difficult for a
researcher looking at either photograph to stumble on the other photograph
without sifting through tens of thousands of images. The embedding neighbors
solves this problem by linking the two related but distinct terms used to
describe the scenes. Both rows together reveal the wide scope of the FSA-OWI
corpus and the broad mandate given to the photographers. The DualNeighbors
algorithm, therefore, illuminates connections that would be hidden by
previous word-based search and recommender systems.

\subsection{News Twitter Reports}

Our second corpus is taken from Twitter, consisting of tweets by news
organizations in the year 2017 \citep{littman2017news}. We compare the
center-left British daily newspaper \textit{The Guardian} and the center-right
daily French newspaper \textit{Le Figaro}. Twenty thousand
tweets were randomly selected from each newspaper, after removing retweets and
anything whose content was empty after removing hashtags and links. We used a
French parser and word embedding to work with the data from \textit{Le Figaro}
and an English parser and embedding to process \textit{The Guardian} headlines
\citep{straka2016udpipe}.

\begin{table*}
\centering
\begin{tabular}{c|cc|ccccc|ccccc}
&&& \multicolumn{5}{|c|}{FSA-OWI} & \multicolumn{5}{|c}{Twitter}   \\
  &$N_{w}$ & $N_{e}$ & $\lambda_2$ & u.c. & dist & $d_{0.9}^{in}$ & $\text{ego}_{0.1}$
  & $\lambda_2$ & u.c. & dist & $d_{0.9}^{in}$ & $\text{ego}^{(3)}_{0.1}$ \\
  \hline
  \hline
& 12 & 0  & 0.002 & 25.1\% &  9.8 & 27 & 17   & $\cdot$ & 57.6\% & 7.3 & 25 & 16   \\
\hline
\parbox[t]{2mm}{\multirow{6}{*}{\rotatebox[origin=c]{90}{Q. Replacement}}}
& 11 & 1  & 0.011 & 15.7\% &  8.4 & 26 & 77   & 0.028   & 11.1\% & 7.3 & 26 & 84   \\
& 10 & 2  & 0.023 & 15.5\% &  8.1 & 25 & 124  & 0.046   & 11.0\% & 7.2 & 27 & 110  \\
& 9  & 3  & 0.038 & 16.3\% &  7.9 & 24 & 158  & 0.056   & 12.2\% & 7.1 & 28 & 129  \\
& 8  & 4  & 0.047 & 17.8\% &  7.8 & 23 & 189  & 0.070   & 14.6\% & 7.0 & 29 & 134  \\
& 7  & 5  & 0.056 & 20.4\% &  7.8 & 22 & 217  & 0.077   & 17.0\% & 7.0 & 29 & 139  \\
& 6  & 6  & 0.061 & 23.8\% &  7.8 & 20 & 238  & 0.085   & 20.6\% & 7.0 & 30 & 137  \\
\hline
\parbox[t]{2mm}{\multirow{6}{*}{\rotatebox[origin=c]{90}{Q. Expansion}}}
& 11 & 1  & 0.002 & 26.8\% &  9.2 & 26 & 50   & 0.028   & 21.0\% & 8.2  & 25 & 61   \\
& 10 & 2  & 0.002 & 31.7\% &  9.3 & 25 & 53   & 0.024   & 31.7\% & 8.9  & 25 & 68  \\
& 9  & 3  & 0.002 & 35.5\% &  9.6 & 24 & 56   & 0.020   & 42.5\% & 10.2 & 26 & 65  \\
& 8  & 4  & 0.002 & 40.8\% &  9.8 & 22 & 59   & 0.010   & 54.2\% & 15.8 & 26 & 61  \\
& 7  & 5  & 0.003 & 47.0\% & 10.8 & 21 & 62   & 0.013   & 59.9\% & 2.3  & 26 & 56  \\
& 6  & 6  & 0.004 & 52.9\% & 10.4 & 20 & 64   & 0.014   & 60.3\% & 1.5  & 25 & 51  \\
\end{tabular}
\caption{Connectivity metrics for similarity graphs. All examples relate each
item to twelve neighbors, with $N_{w}$ word neighbors and $N_{e}$ embedding
neighbors.
For comparison, we show the results using both query replacement (as
described in the DualNeighbors algorithm) and with the query expansion
method suggested in the papers discussed in Section~\ref{ssec:rec}.
The metrics give the (undirected) spectral
gap $\lambda_2$, the proportion of directed pairs of items that are unconnected
across directed edges (u.c.), the average distance (dist) between connected
pairs of items, the 90th percentile of the in-degree ($d_{0.9}^{in}$), and
the 10th percentile of the number of neighbors within three links
($\text{ego}^{(3)}_{0.1}$).}
\label{tab:metrics}
\end{table*}

In Table~\ref{tab:captions} we see two examples of the word and embedding
nearest neighbors.
The first tweet shows how the English word
``shareholders'' is linked both to its closest direct translation
(``actionnaires'') as well as the more generic ``investisseur''.
In the next example the
embedding links the search term to its most direct translation. ``Red
carpet'' becomes ``tapis rouge''.  Once translated,
we see that the themes linked to by both newspapers are similar, illustrating
the algorithm's ability to traverse linguistic boundaries within a corpus.
Joining headlines across these two newspapers, and by extension the longer
articles linked to in each tweet, makes it possible to compare the
coverage of similar events across national, linguistic, and ideological
boundaries. The connections shown in these two examples were only found through
the use of the implicit translations given by the multilingual word embeddings
as implemented in the DualNeighbors algorithm.

\section{Quantitative Evaluation} \label{sec:quant}

\subsection{Connectivity}

We can study the set of recommendations given by our algorithm as a
network structure between documents in a corpus. This is useful because there
are many established metrics measuring the degree of connectivity within a
given network. We will use five metrics to understand the network structure
induced by our algorithm: (i) the algebraic connectivity, a measurement of
whether the network has any bottlenecks \citep{fiedler1973algebraic},
(ii) the proportion of document pairs
that can be reached using edges, (iii) the average minimum distance between
connected pairs of documents, (iv) the distribution of in-degrees, the number
of other documents linking into a given document \citep{even1975network},
and (v) the distribution of
third-degree ego scores, the number of documents that can be reached by moving
along three or fewer edges \citep{everett2005ego}.
The algebraic connectivity is defined over an
undirected network; the other metrics take the direction of the edge into
account.

Table~\ref{tab:metrics} shows the five connectivity metrics for various choices
of $N_w$ and $N_e$. All of the examples use a total of $12$ recommendations
for consistency. Generally, we see that adding more edges from the
(query expansion) word
embedding matrix produces a network with a larger algebraic connectivity,
lower average distance between document pairs, and larger third-degree ego
scores. The distribution of in-degrees becomes increasingly variable, however,
as more edges get mapped to a small set of hubs (documents linked to from a
very large number of other documents). These two effects combine so that the
most connected network using both corpora have $10$ edges from word
similarities and $2$ edges from the word embedding logic. Generally, including
at least one word embedding edges makes the network significantly more
connected. The hubness of the network slowly becomes an issue as the proportion
of embedding edges grows relative to the total number of edges.

To illustrate the importance of using query replacement in the word
embedding neighbor function, the table also compares our approach (query
replacement) to that
of query expansion. That is, what happens if we retain the original term in
the  embedding neighbor function $f$, as used in Equation~\ref{expcorp},
rather than replacing it. Table~\ref{tab:metrics} shows that the query
replacement approach of the DualNeighbors algorithm provides a greater degree
of connectivity across all five metrics and any number of embedding neighbors
$N_e$. Therefore, this modification serves as an important contribution and
distinguishing feature of our approach.

\subsection{Relevance} \label{sec:rel}

It is far more difficult to quantitatively assess how relevant the
recommendations made by our algorithm are to the starting document. The degree
to which an item is relevant is subjective. Also, our goal is to find links
across the corpus that share thematic similarities but also cut across
languages and discourses, so a perfect degree of similarity between
recommendations is not necessarily ideal. In order to make a quantitative
assessment of relevancy, we constructed a dataset of $3,000$ randomly collected
links between documents from each of our two corpora. We hand-labelled whether
or not the link appeared to be `valid'. This was done according to whether the
links between any of the terms used to link the two texts together used the
terms in the same word sense. For example, we flagged as an invalid connection
a link between the word ``scab'' used to describe a skin disease and ``scab''
as a synonym for strikebreaker. While a link being `valid' does not
guarantee that there will be an interesting connection between two documents,
it does give a relatively unambiguous way of measuring whether the links found
are erroneous or potentially interesting.

\begin{table}
\centering
\begin{tabular}{c|cc|cc}
& \multicolumn{2}{|c|}{FSA-OWI} & \multicolumn{2}{|c}{Twitter} \\
Pos. & TF-IDF & Emb. & TF-IDF & Emb. \\
  \hline
  \hline
1-3  & 0.88\% & 2.66\% & 6.34\% &  9.52\% \\
4-8  & 1.27\% & 2.54\% & 9.09\% & 10.32\% \\
9-12 & 5.17\% & 3.16\% & 9.40\% & 13.55\%
\end{tabular}
\caption{Taking a random sample of 3000 links from each corpus, the proportion
of links between terms that were hand-coded as `invalid' organized by corpus,
neighbor type, and the position of the link in the list of edges. See
Section~\ref{sec:rel} for the methodology used to determine validity.}
\label{tab:relevance}
\end{table}

The results of our hand-tagged dataset are given in Table~\ref{tab:relevance},
with the proportion of invalid links grouped by corpus, edge type, and
the position of the edge within the list of possible nearest neighbors.
Overall, we see that the proportion of valid embedding neighbors is nearly as
high as the word neighbors across both corpora and the number of selected
neighbors. This is impressive because there are many more ways that the word
embedding neighbors can lead to invalid results. The results of
Table~\ref{tab:relevance} illustrate, however, that the embedding neighbors
tend to find valid links that use both the source and target words in the same
word sense. This is strong evidence that the DualNeighbors algorithm increases
the connectivity of the recommendations through meaningful cross-discursive and
multilingual links across a corpus.


\section{Implementation} \label{sec:imp}

To facilitate the usage of our method in the exploration of textual data, we
provide an open-source implementation of the algorithm in the R
package \textbf{cdexplo}.\footnote{The package can be downloaded and installed
from \url{https://github.com/statsmaths/cdexplo}}
The package takes raw text as an input and produces an interactive website
that can be used locally on a user's computer; it therefore requires only
minimal knowledge of the R programming language. For example, if a corpus is
stored as a CSV file with the text in the first column, we can run the
following code to apply the algorithm with $N_w$ equal to $10$ and $N_{e}$
equal to $2$:
\begin{Verbatim}[fontsize=\small]
library(cdexplo)
data <- read.csv("input.csv")
anno <- cde_annotate(data)
link <- cde_dual_neigh(anno, nw = 10, ne = 2)
cde_make_page(link, "output_location")
\end{Verbatim}
The source language and presence of metadata, including possible image URLs,
will be automatically determined from the input, but can also be manually
specified. The image in Figure~\ref{fig:embed} is a screen-shot from the output
of the package applied to the FSA-OWI caption corpus.

\section{Conclusions}

We have derived the DualNeighbors algorithm to assist in the exploration of
textual datasets. Qualitative and quantitative analyses
have  illustrated how the algorithm cuts across linguistic boundaries
and improves the connectivity of the recommendation algorithm without
a significant decrease to the relevancy of the returned results.

Language is impacted by cultural factors surrounding the writer
and their subject. Syntactic and lexical choices serve as strong
signals of class, race, education, and gender. The ability to connect and
transcend the boundaries constructed by language while exploring textural data
offers a powerful new approach to the study of cultural datasets.
Our open-source implementation assists in the application of the DualNeighbors
approach to new corpora. Furthermore, the computed recommendations can be
directly adapted as a recommendation algorithm for digital public projects,
allowing the exploratory benefits afforded by our technique to be available to
a wider audience.

One avenue for extending the DualNeighbors algorithm is to further refine the
process of constructing a lexicon and corresponding word embedding. Most of the
errors we detected in the experiment in Section~\ref{sec:rel} were the result
of proper nouns and noun phrases that do not make sense when embedding each
individual word. Recent work has shown that better pre-processing can alleviate
some of these difficulties \cite{trask2015sense2vec}. We also noticed,
particularly over the jargon-heavy Twitter news corpus, that many key phrases
were missing from our embedding mapping. Research on sub-word
\cite{bojanowski2017enriching} and character level embeddings
\cite{santos2014learning, zhang2015character} could be used to address terms
that are outside of the specified lexicon.

\section*{Acknowledgements}

We thank an anonymous reviewer whose comments suggested an additional
motivation for our work. These suggestions have been incorporated into the
final version of the paper.

\bibliographystyle{acl_natbib}
\bibliography{acl2018}

\end{document}